\documentclass[twocolumn]{el-author}

\newcommand{\ua}{\uparrow}
\newcommand{\nc}{\newcommand}
\nc{\da}{\downarrow} \nc{\hc}{\hat{c}} \nc{\hS}{\hat{S}}
\nc{\bra}{\langle} \nc{\ket}{\rangle} \nc{\eq}{equation (\ref}
\nc{\h}{\hat} \nc{\hT}{\h{T}}\nc{\be}{\begin{eqnarray}}
\nc{\ee}{\end{eqnarray}}\nc{\rd}{\textrm{d}}\nc{\e}{eqnarray}\nc{\hR}{\hat{R}}\nc{\Tr}{\mathrm{Tr}}
\nc{\tS}{\tilde{S}}\nc{\tr}{\mathrm{tr}}\nc{\8}{\infty}\nc{\lgs}{\bra\ua,\phi|}\nc{\rgs}{|\ua,\phi\ket}
\nc{\hU}{\hat{U}}\nc{\lfs}{\bra\phi|}\nc{\rfs}{|\phi\ket}\nc{\hZ}{\hat{Z}}\nc{\hd}{\hat{d}}\nc{\mD}{\mathcal{D}}
\nc{\bd}{\bar{d}}\nc{\bc}{\bar{c}}\nc{\mc}{\mathcal}\nc{\ea}{eqnarray}\nc{\mG}{\mathcal{G}}\nc{\bce}{\begin{center}}
\nc{\ece}{\end{center}}
\date{7th July 2019}

\begin{document}

\title{Appearance invariant Entry-Exit matching using visual soft biometric traits}

\author{Vinay Kumar V and P. Nagabhushan}

\abstract{The problem of appearance invariant subject recognition for Entry-Exit surveillance applications is addressed. A novel Semantic Entry-Exit matching model that makes use of ancillary information about subjects such as height, build, complexion and clothing color to endorse exit of every subject who had entered private area is proposed in this paper. The proposed method is robust to variations in  clothing. Each describing attribute is given equal weight while computing the matching score and hence the proposed model achieves high rank-k accuracy on benchmark datasets. The soft biometric traits used as a combination though cannot achieve high rank-1 accuracy, it helps to narrow down the search to match using reliable biometric traits such as gait and face whose learning and matching time is costlier when compared to the visual soft biometrics.}

\maketitle

\section{Introduction}
Intelligent video surveillance systems overcame the limitations in human ability to diligently watch and monitor multiple live video surveillance footages\cite{1}. Intelligent video surveillance domain witnessed extensive research in the past two decades thus extending its applications to intruder detection and validation, crime prevention, elderly people and children monitoring. Today, public places such as shopping malls, airports, bus and rail stations are completely under surveillance ambit except in few areas such as toilets and changing rooms which are referred to as \textit{private areas} where installing surveillance cameras is considered breach of privacy. This is often seen as a hindrance for security systems in crime prevention and public safety. As a solution, the notion of Entry-Exit surveillance (EES) \cite{2} deals with monitoring of subjects entering and exiting private areas. The key objective is to assure that the subjects who had entered private areas exit in time without much variations in their appearances that may lead to suspicion. Every subject who enter private area is labeled and saved in gallery and every subject who exit the private area is considered as probe and has to be matched with subjects available in the gallery. 
\subsection{Relevance of the problem}
The problem of entry-exit matching can be related to person re-identification problem where the aim is to match subjects moving between surveillance ambits of non-overlapping cameras. For every probe subject, the matching subject in the gallery set should have high matching rank when compared with other subjects. Detailed survey on person re-identification can be found in \cite{3,4}. In conventional person re-identification systems, it is assumed that the subject is to be matched across camera views on the same day and hence the issue of variation in appearance of subjects due to change in clothing is under addressed. However, in entry-exit surveillance, the temporal gap between entry and exit where the subjects move out of surveillance ambit envisage the possibility of change in appearance of subjects with respect to clothing, carrying and head masking. Traces of appearance invariant subject re-identification solutions available in the literature include analysis of gait and motion patterns in \cite{5}. However, state of the art gait recognition algorithms suffer due to variations in the walking directions. Face biometric trait is another promising attribute for subject re-identification that can be robust to changes in clothing and carrying conditions. However, capturing of face attributes is limited to very few frames due to the distance of subjects from the camera. Adding to it, the possibility of face occlusions due to subjects overlapping as well as religious and cultural practices in unconstrained environments limits extraction of face attributes from surveillance videos. 
\par On the other hand, visual soft biometric attributes such as clothing color, height, body-build, accessories possessed by the subjects can be extracted from low resolution video frames. Most of the current state of the art person re-identification methods mainly focus on clothing attributes for matching. However, in entry-exit surveillance, due to possibility of variations in clothing of subjects, relatively less importance must be attributed for clothing color but cannot be completely dropped as it has high discriminating ability in majority of the cases. The probability of change in clothing in places such as toilets and baby feeding rooms is relatively lower when compared to that of dress changing rooms in clothing outlets. Hence, it is necessary to analyze the reliability of clothing color in different scenarios. Height of the subjects is most view invariant as reported in \cite{6} and capturing of the attribute during subjects appearance in a predefined region of interest in the camera view scene (entrances of private areas in Entry-Exit surveillance scenario),  makes it reliable. Similarly, build of the subjects can be captured by computing the height to width ratio provided the video footages are captured from a still camera with no variations in the view angle. Unlike height attribute, build attribute is pose variant. However, computing the vertical projections of the silhouettes of the segmented subjects' bounding boxes makes it discriminative. Lastly, skin complexion can be another promising attribute if illumination variations in the multiple camera views is handled. 
\par The above discussed soft biometrics are unique in nature and are to be given equal weight for their discriminative ability under different scenarios. Also, as one soft biometric attribute cannot single handedly identify subjects due to intra-class variations and inter class similarities, combinations of these attributes to an extent can predict the subjects by matching them with the gallery samples.

\subsection{Motivations and Contributions}
With the above discussion, it can be inferred that though soft biometrics are not completely reliable to recognize subjects, they provide prominent clues and help in narrowing down the search. Also, learning of soft biometrics is computationally faster as compared to the classical visual biometrics such as face and gait. This motivates us to explore the efficiency of soft biometrics in Entry-Exit matching.

\par The innovations accomplished in this paper can be summarized as follows:
\begin{itemize}
	\item First, a set of hybrid features representation that is robust to possible appearance variations is introduced.
	\item An ensemble based approach for handling heterogeneous matching results from individual soft biometrics.
	\item Matching analysis based on single-camera as well as two-camera based Entry-Exit surveillance model.
		
\end{itemize}
\par The proposed method is evaluated using EnEx dataset that comprise of Entry-Exit surveillance data using Single-camera and also with our own database that also comprise of Entry-Exit surveillance data but using two field-of-view-overlapping cameras.


\section{Statement of the problem}
Given the input images of subjects that are segmented from the surveillance video frames, the aim is to represent each subject with a set of highly discriminative soft biometric features such as clothing color, height, body-build and complexion. Features from the subjects that are classified to have entered private areas are extracted and are saved in gallery with labels. For every subject who exit from private area, features are extracted and are matched with samples in gallery and the matching score is computed with each labeled subject. Certainly, 100\% classification accuracy cannot be expected from rank-1 but from rank-k predictions and the goal is to find the value of k with which the search can be narrowed down from n to k where k<<n.

\section{Proposed Model}
The proposed subject recognition model is outlined in two modules as described below.
\begin{itemize}

\item Feature representation: A novel feature representation that comprise of set of features that include clothing color as well as subjects height, body-build and complexion is introduced thus making the model robust to clothing changes. Detailed discussion on each individual feature types can be found in the next subsection.
\item Learning and Recognition: Each feature type is analyzed for its discriminative and correlating abilities on inter-class and intra-class samples respectively and the transformation function that maximizes the inter-class separability and intra-class associativity is computed.
Subject recognition is performed based on collective confidence of the visual soft biometrics ensembles.
\end{itemize}
\subsection{Feature Representation}
A set of heterogeneous features represent individual subject as discussed below. 
\subsubsection{Clothing Color}

The given input image of the subject is decomposed into head, torso and leg regions based on \cite{7}. Torso and leg regions are individually analyzed. Low level features such as color and texture used in conventional person re-identification methods exhibit high discriminative ability on different camera views. The computational complexity of color features is faster as compared to texture features \cite{8} and hence the proposed model is confined to color features. \par RGB color space is robust to variations in translation, rotation and scaling factors but suffers from temporal light changes outdoor throughout the day as well as illumination variations due to different camera views.  
HSV color space is interpreted as color space commonly perceived by humans and is considered efficient enough for color based analysis, the RGB to HSV color transformation is time consuming and slows down learning. 
On the other hand, YCbCr color space clearly differentiate the luminance and color components. The Y component corresponds to luminance and Cb and Cr components correspond to chrominance where Cb is  
blue-difference chrominance and Cr is red-difference chrominance and the computational complexity of RGB to YCbCr tranformation is of the order \textit{h X w} where h and w are height and width of the input subject image. 
Histograms of Cb and Cr components of the torso and leg regions are computed for each subject and are concatenated to a single feature vector.
\subsubsection{Height}
Height is one of the prominent features that helps in narrowing down the search by clustering subjects of similar heights. Height of the subject is captured with reference to the entrance of private areas. As the features of the subjects who enter private areas is saved in the gallery only on confirmation of his/her entry, learning is done in time. Also, during exit, the subjects have to cross the entrance first and hence, height is captured with other set of features. As the height feature captured is the relative height and not the actual height, it is normalized to the values between 0 and 1.

\subsubsection{Body-build}
Extracting height feature near to the entrances of private areas also provide scope for extracting build of the subject by computing the maximum height to width ratio of the bounding boxes of the subjects when they cross the entrance. Variations in pose of the subject is the major challenge to be addressed. Vertical projection profile is computed for the bounding boxes and threshold t is determined with experiments to eliminate hand and leg swings thus segmenting the subject image based on torso distribution.

\subsubsection{Skin Complexion}  
Skin complexion is another important visual attribute that helps in grouping subjects with same complexion. The segmented head region is analyzed for skin components using YCbCr color model with the threshold for skin detection as reported in \cite{9}. Skin region is segmented from front, lateral and oblique views except from back view.

\subsection{Matching}
For each feature type, Linear Discriminant analysis (LDA) is applied for data disassociation, by  projecting the feature matrix onto a subspace that maximizes the ratio of inter-class to intra-class distribution, using the Fisher's criterion.

For every class $ C_{i} $, the separability $ d_{i} $ between samples  $s \in C_{i}$ is computed using 
\begin{equation}\label{equation1}
d_{i} = \sum{(s-\bar{s_{i}})(s-\bar{s_{i}})'}
\end{equation} 
where $ \bar{s_{i}} $ is the mean of the class $ C_{i} $\\
The intra-class separability matrix d for $ n $ classes is given by 
\begin{equation}\label{equation2}
d = \sum_{i=1}^{n}{d_{i}}
\end{equation} 
and the intra-class compactness is given by
\begin{equation}\label{equation3}
Q = 1/d
\end{equation} 

The disassociativity matrix D between classes is computed using

\begin{equation}\label{equation4}
D = \sum_{i=1}^{n}{m_{i}(\bar{s_{i}} - \bar{s})(\bar{s_{i}} - \bar{s})'}
\end{equation} 
where $ m_{i} $ is the number of training samples in class $ C_{i} $ and $ \bar{s} $ is the overall mean.

The transition matrix $ T $ that maximizes the spread between the classes and compactness within the classes is given by

\begin{equation}\label{equation5}
T = |D| |Q|
\end{equation}

So, given a probe subject $ p $, the features are extracted and every soft biometric is operated with transition function in equation \ref{equation5} and the classification is performed for every soft biometric by computing the euclidean distance of the probe with every gallery sample and thus all the gallery classes are ranked for every soft biometric. Subject recognition is done based on the collective voting decision among the different soft biometrics ensemble thus giving equal prominence for each soft biometric . 
Confidence $ Cf $ is computed for every gallery class $ C_{i} $ for each soft biometric $ f $ using
\begin{equation}\label{equation6}
Cf(C_{i},f) = \dfrac{n - rank(C_{i},f) + 1}{n}
\end{equation}

Hence, the Collective confidence $ CF $ of the model for each gallery class $ C_{i} $ is

\begin{equation}\label{equation7}
CF(C_{i}) = \sum_{j=1}^f{\dfrac{Cf(C_{i},j)}{f}}
\end{equation}

Rank for every gallery class is assigned based on the collective confidence calculated using equation \ref{equation7}. 
\section{Experiments and Discussions}
The proposed method is tested using EnEx dataset as well as our own database. EnEx dataset provide EES data using single camera and our database provide EES data using two cameras. However, our database can be used for single camera based analysis as well. Firstly, with single camera view where the camera is placed so as to view the entrance of private areas without intruding the privacy of the public. Here, the camera captures flipped views of subjects during entry and exit where generally back view of the subject is visible during entry and front view during exit or if the camera is placed so as to have lateral view of the subject, then entry-exit  shall have left-right flipped view variations. Next, with two view-overlapping cameras so as to have $360^{o}$ view of the subject where one camera compliment the other by having flipped view of the subject simultaneously. 

\subsection{Experimental setup}

\begin{figure}[htbp]
	\centering
		\includegraphics[height=0.30\linewidth]{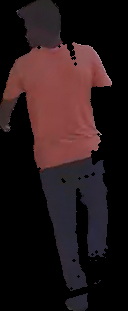}
	\includegraphics[height=0.30\linewidth]{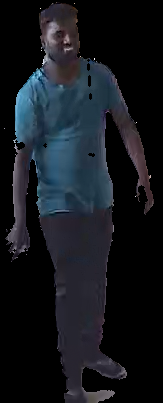}
	\caption{Sample Entry-Exit image pair with variation in appearance from our database using Single Camera}
	\label{fig:SingleCamera}
	
\end{figure}

\begin{figure}[htbp]
	\centering
	\includegraphics[height=0.30\linewidth]{Camera2_Entry}
	\includegraphics[height=0.30\linewidth]{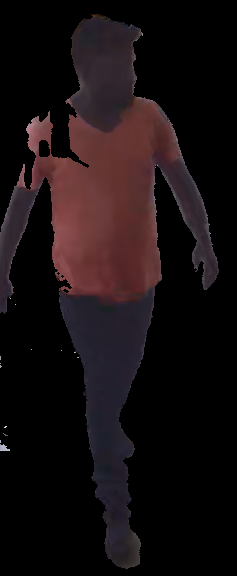}
	
	\includegraphics[height=0.30\linewidth]{Camera2_Exit1}
	\includegraphics[height=0.30\linewidth]{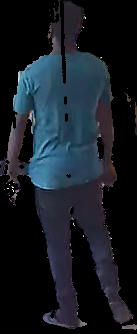}
	\caption{Sample Entry-Exit image pair with variation in appearance from our database captured using two cameras}
	
	\label{fig:TwoCameras}
	
\end{figure}

The input to the system is the subjects images extracted from video frames of the datasets using \cite{10} deeply learned people detection method. The frames are background subtracted before applying the people detectors. Two sets of images - gallery and probes. Gallery set contain images captured while subjects entered whereas probe set contained images captured while subjects exited. Figure \ref{fig:SingleCamera} shows sample images of subjects in gallery and probe using single camera and figure \ref{fig:TwoCameras} shows sample images of subjects in gallery and probe using two cameras. The image pair in first row shows Entry of the subject captured in two different cameras and the second row shows subject exiting the private area 

The height and body-build attributes are captured when the subject crossed the entrances of private areas and other attributes such as clothing color and skin complexion were extracted after normalizing the input images to standard size of 128 $ \times $ 64. Then the input images are decomposed into head, torso and leg regions and these are converted to YCbCr color model and histograms of Cb and Cr components are extracted with 24 bins per channel for each region of a subject. 
Skin regions are extracted from each input image and the mean of Cb and Cr components are computed. 

Initially simulations were carried out on EnEx dataset, here the size of the private area, that is gallery was assumed to be 10, 25 and 50 respectively. The gallery contained 30 training images for each subject among which 5 images provided height and body-build metrics. Hence, the height and body-build attributes are trained separately. For every probe subject, single image that contained all the attributes was selected to match with all the samples in the gallery. Later, the simulations were extended to our own dataset where pair of images for clothing and skin complexion attributes are provided for training and testing and the features are concatenated to a single vector for representation. 

The notion of Entry-Exit surveillance is novel and the proposed model is first of its kind that solved Entry-Exit matching which involved appearance invariant person re identification. However, comparative analysis is provided with \cite{11} and \cite{12} methods by evaluating them on the EnEx and our own database. Tables \ref{tab:EnEx25},\ref{tab:EnEx50}, \ref{tab:EnEx225} and \ref{tab:EnEx250} provide matching scores of the proposed method in comparison with \cite{11} and \cite{12}. 

\subsection{Discussion}
With the results it is evident that the state of the art person re identification systems suffer due to variations in the clothing of the subjects and hence demand for active research in appearance invariant person re identification. The proposed model though is robust to clothing variations, it suffers due to uniformity in height, build and complexion of subjects of same race. Also, extraction of skin attributes from subjects who cover entire body with religious attires is challenging. The height attribute though looks promising, it suffers due to variations in head accessories of the subjects. Overall, the evaluation results of the proposed model on the EES specific datasets can be used as benchmark by the research community to compare their works on the EES matching problem using visual soft biometrics.
\begin{table}
		\centering
	\begin{tabular}{ |p{3.0cm}||p{1cm}|p{1cm}|p{1cm}|  }
		\hline
		\multicolumn{4}{|c|}{Gallery size = 25} \\
		\hline
		Rank & 1 & 5 &10\\
		\hline
		RS-KISS   & 0.079    &0.316&   0.486\\
		Ensemble Learning&   0.094  & 0.367   &0.491\\
		Proposed model &0.231 & 0.489&  0.867\\
		\hline
	\end{tabular}
\caption{Matching rates on EnEx dataset, gallery size = 25}
\label{tab:EnEx25}
\end{table}

\begin{table}
	\centering
	\begin{tabular}{ |p{3.0cm}||p{1cm}|p{1cm}|p{1cm}|  }
		\hline
		\multicolumn{4}{|c|}{Gallery size = 50} \\
		\hline
		Rank & 1 & 5 &10\\
		\hline
		RS-KISS   & 0.071    &0.283&   0.423\\
		Ensemble Learning&   0.082  & 0.326   &0.463\\
		Proposed model &0.217 & 0.466&  0.812\\
		\hline
	\end{tabular}
	\caption{Matching rates on EnEx dataset, gallery size = 50}
	\label{tab:EnEx50}
\end{table}

\begin{table}
\centering	
	\begin{tabular}{ |p{3.0cm}||p{1cm}|p{1cm}|p{1cm}|  }
		\hline
		\multicolumn{4}{|c|}{Gallery size = 25} \\
		\hline
		Rank & 1 & 5 &10\\
		\hline
		RS-KISS   & 0.091    &0.352&   0.533\\
		Ensemble Learning&   0.162  & 0.431   &0.616\\
		Proposed model &0.366 & 0.581&  0.891\\
		\hline
	\end{tabular}
	\caption{Matching rates on our dataset, gallery size = 25}
		\label{tab:EnEx225}
\end{table}

\begin{table}
	\centering
	\begin{tabular}{ |p{3.0cm}||p{1cm}|p{1cm}|p{1cm}|  }
		\hline
		\multicolumn{4}{|c|}{Gallery size = 50} \\
		\hline
		Rank & 1 & 5 &10\\
		\hline
		RS-KISS   & 0.088    &0.263&   0.411\\
		Ensemble Learning&   0.113  & 0.412   &0.493\\
		Proposed model &0.342 & 0.563&  0.838\\
		\hline
	\end{tabular}
	\caption{Matching rates on our dataset, gallery size = 50}
		\label{tab:EnEx250}
\end{table}

\section{Conclusion}
With this paper, it can be inferred that in real-time tracking it is important to narrow down the search based on predictions using visual attributes whose learning and recognition is faster than the reliable attributes whose learning and recognition rates are higher and faster with the narrow galleries. The proposed novel Entry-Exit subject matching method show good rank-10 accuracy on the standard datasets and thus enkindle competitive research in Entry-Exit surveillance domain. \vskip3pt
\ack{This work has been supported by The University Grants Commission, India }

\vskip5pt

\noindent Vinay Kumar V (\textit{Department of studies in Computer Science, Univeristy of Mysore, Mysuru, India}) and P Nagabhushan (\textit{Indian Institute of Information Technology-Allahabad, Prayagraj, India})
\vskip3pt

\noindent E-mail: vkumar.vinay@ymail.com

\end{document}